%% file: root.tex
\documentclass[lettersize,journal]{IEEEtran}

\usepackage{graphicx}
\usepackage{enumitem}
\usepackage{multirow}
\usepackage{algorithm}
\usepackage{algcompatible}
\usepackage{amsmath}
\usepackage{amssymb}
\usepackage{hhline}
\usepackage{xcolor}

\algnewcommand\algorithmicreturn{\textbf{return}}
\algnewcommand\RETURN{\State \algorithmicreturn}

\usepackage{caption, subcaption}
\definecolor{lbrown}{HTML}{80471C}

\makeatletter
\renewcommand{\COMMENT}[1]{\Statex \hskip\ALG@thistlm   {\color{lbrown}\(\triangleright\)\textit{#1}}}
\makeatother

\usepackage[
	activate   = {true},
	protrusion = false,
	expansion  = true,
	kerning    = true,
	spacing    = true,
	tracking   = false,
	auto       = true,
	selected   = true,
	factor     = 2000,
	stretch    = 50,
	shrink     = 20,
]{microtype}

\usepackage[english=american]{csquotes}
\MakeOuterQuote{"}

\makeatletter
\let\NAT@parse\undefined
\makeatother
\usepackage[pdfa,colorlinks,bookmarksopen,bookmarksnumbered,allcolors=brown]{hyperref}
\usepackage[english]{babel}

\usepackage[nameinlink,capitalise]{cleveref}
\crefname{line}{line}{lines}
\crefname{figure}{Fig.}{Figs.}
\Crefname{figure}{Fig.}{Figs.}
\crefname{equation}{Eq.}{Eqs.}
\Crefname{equation}{Eq.}{Eqs.}
\crefname{section}{Sec.}{Secs.}
\Crefname{section}{Sec.}{Secs.}
\crefname{definition}{Def.}{Defs.}
\Crefname{definition}{Def.}{Defs.}
\crefname{algorithm}{Alg.}{Algs.}
\Crefname{algorithm}{Alg.}{Algs.}
\crefname{assumption}{Asm.}{Asms.}
\Crefname{assumption}{Asm.}{Asms.}
\crefname{theorem}{Thm.}{Thms.}
\Crefname{theorem}{Thm.}{Thms.}
\crefname{table}{Tbl.}{Tbls.}
\Crefname{table}{Tbl.}{Tbls.}
\crefname{subassumption}{Asm.}{Asms.}
\Crefname{subassumption}{Asm.}{Asms.}

\usepackage{tabularx}
\usepackage{booktabs}
\usepackage{colortbl}
\usepackage{overpic}

\definecolor{lgray}{HTML}{F0F0F0}

\newcolumntype{I}{>{\centering}p{0.13\textwidth}}
\newcolumntype{Y}{>{\centering\arraybackslash}X}

\usepackage{amsthm}

\newtheoremstyle{plaindefnobox}
  {}{}             %
  {}               %
  {}               %
  {\bfseries}      %
  {:}              %
  { }              %
  {\thmname{#1} \thmnumber{#2}: \thmnote{\em#3}} %
\theoremstyle{plaindefnobox}

\begin{document}
\title{\LARGE \bf Underwater Multi-Robot Simulation and Motion Planning in \emph{Angler}}

\author{Akshaya Agrawal$^{1}$, Evan Palmer$^{1*}$, Zachary Kingston$^{2}$, and Geoffrey A. Hollinger$^{1}$
\thanks{$^{1}$Collaborative Robotics and Intelligent Systems (CoRIS) Institute, Oregon State University, \texttt{\{agrawaak, palmeeva, geoff.hollinger\}@oregonstate.edu}}
\thanks{$^{2}$Department of Computer Science, Purdue University \texttt{zkingston@purdue.edu}}
\thanks{This work was supported in part by ONR awards N0014-21-1-2052 and N00014-22-1-2114.}
\thanks{The work of Evan Palmer was supported by the National Defense Science and Engineering Graduate (NDSEG) Fellowship.}
}

\maketitle
\input{files/00-abstract}


\input{files/01-introduction}

\input{files/02-related-work}
\input{files/03-simulation-features}
\input{files/04-multi-robot-motion-planning}
\input{files/05-conclusion-and-future-works}

\bibliographystyle{IEEEtran}
\bibliography{bibliography}

\end{document}

%% file: files/00-abstract.tex
\begin{abstract}

Deploying multi-robot systems in underwater environments is expensive and lengthy; testing algorithms and software in simulation improves development by decoupling software and hardware. 
However, this requires a simulation framework that closely resembles the real-world.
\emph{Angler} is an open-source framework that simulates low-level communication protocols for an onboard autopilot, such as \emph{ArduSub}, providing a framework that is close to reality, but unfortunately lacking support for simulating multiple robots. We present an extension to \emph{Angler} that supports multi-robot simulation and motion planning. Our extension has a modular architecture that creates non-conflicting communication channels between Gazebo, ArduSub Software-in-the-Loop (SITL), and MAVROS to operate multiple robots simultaneously in the same environment. Our multi-robot motion planning module interfaces with cascaded controllers via a \texttt{JointTrajectory} controller in ROS~2. We also provide an integration with the Open Motion Planning Library (OMPL), a collision avoidance module, and tools for procedural environment generation. 
Our work enables the development and benchmarking of underwater multi-robot motion planning in dynamic environments.

\end{abstract}

%% file: files/01-introduction.tex
\section{Introduction}\label{section:introduction}

Underwater multi-robot teams are capable of addressing complex, collaborative tasks such as connecting pipes in offshore sub-sea pipeline networks, and can explore areas faster than one robot alone~\cite{roger2021twinbot, marani2009underwater,sanz2010trident}. However, deploying multi-robot systems underwater is risky, difficult, requires significant hardware and labor investment, and is time-intensive. Many components of the system (e.g., integration of different modules such as sensors, controllers, and planners) can be evaluated in simulation to reduce these costs, and high-level integrations can be tested before deployment on real hardware. 

\begin{figure}[t]
  \centering
  \includegraphics[width=9cm,trim={0cm 0cm 0cm 0cm},clip]{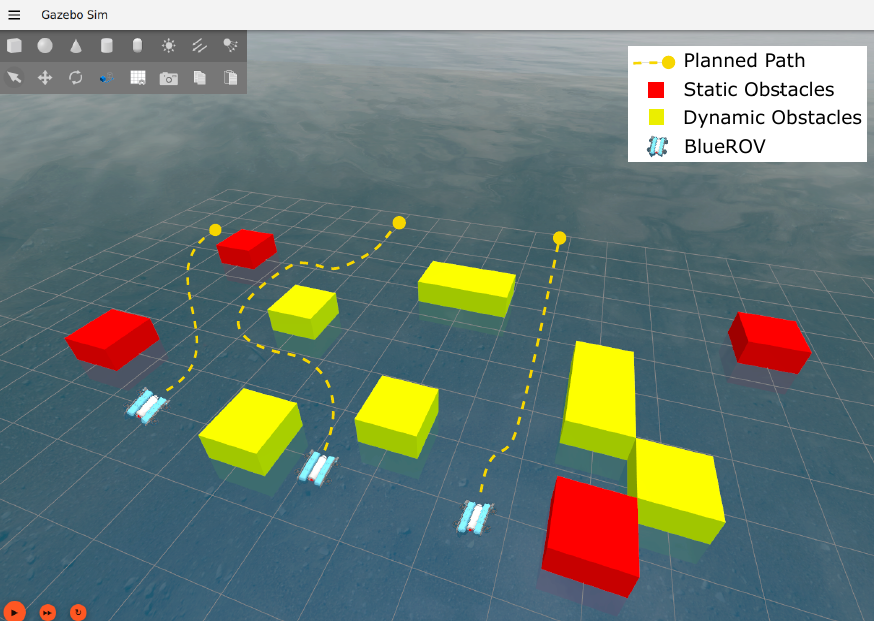}
  \caption{A scene in \emph{Angler} showcasing our multi-robot extension, which simulates a team of robots influenced by ocean currents. 
  These robots are tasked with autonomously navigating around obstacles in a collaborative exploration mission. 
  Our multi-robot extension to \emph{Angler} establishes conflict-free interfaces between Gazebo, ArduSub SITL, and MAVROS, enabling coordinated or individual control of each robot. 
  Additionally, we provide tools for integrating controllers with the motion planning through OMPL and for generating environments that include both static and dynamic obstacles.}
  \label{figure:simulator}%
\end{figure} 

In particular, motion planning is a core capability of autonomous underwater robots as it enables collision-free navigation. Fundamental to motion planning is collision detection, a computationally expensive building block which increases in complexity with the team size of robots. In the underwater case, we are also interested in dynamic environments where objects are affected by ocean currents and buoyancy. As objects in the environment are moving, online motion planning is required to maintain safety; this requires the pose of all the robots and obstacles. A fast, dynamic collision detection module is required for online multi-robot motion planning.

Several simulation frameworks address subsets of these requirements for underwater robotics~\cite{grimaldi2025stonefish, edward2025underwater, gezer2022working, palomeras2012cola2}. \emph{Angler}\footnote{\url{https://github.com/Robotic-Decision-Making-Lab/blue}\label{footnote:blue}}{\color{brown}\textsuperscript{,}}\footnote{\url{https://github.com/Robotic-Decision-Making-Lab/angler}} is an open source software framework that, unlike other simulators, interfaces with ArduSub software-in-the-loop (SITL)~\cite{ArduSub}, enabling testing of software in simulation while maintaining a close-to-hardware deployment setup, enabling eventual deployment on real hardware. ArduSub SITL provides close to real-world hardware-software integration pipeline by simulating a flight control system of a real underwater robot. \emph{Angler} is based on the latest Gazebo for simulation and is built using ROS~2. Although this framework is specifically designed for BlueROV2 Heavy\footnote{\url{https://bluerobotics.com/}} and BlueROV2 Heavy mounted with Reach Alpha 5 Manipulator\footnote{\url{https://reachrobotics.com/}}, it can be adapted to other underwater vehicles. Despite these strengths, \emph{Angler} does not provide support for motion planning out of the box and also lacks support for multi-robot systems due to difficulties in setting up multiple, modular ROS~2, ArduSub SITL, and MAVROS communication channels for every robot in a multi-robot system, while also supporting coordinated control.

This work extends \emph{Angler} to support multi-robot simulation and motion planning, shown in ~\cref{figure:simulator}. 
Our extension provides a platform for both high-level multi-robot coordinated planning approaches and online dynamic replanning. Our code is available in the \emph{Angler}\footref{footnote:blue} repository.
The key contributions of this paper include: 
\begin{enumerate}
    \item  A multi-robot simulator that supports virtual autopilot capabilities through ArduSub SITL for underwater vehicles.
    \item  Integrated tools for motion planning including real-time environment feedback for online planning, collision detection, and an environment generator to support evaluation in diverse environments.
    \item Compatibility with any set of controllers that can be integrated below the \texttt{JointTrajectory} controller from the \texttt{ros2\_control} framework.
\end{enumerate}

%% file: files/02-related-work.tex
\section{Related Work and Background}

\begin{figure}[t]
  \centering
  \includegraphics[width=8cm,trim={0cm 0cm 0cm 0cm},clip]{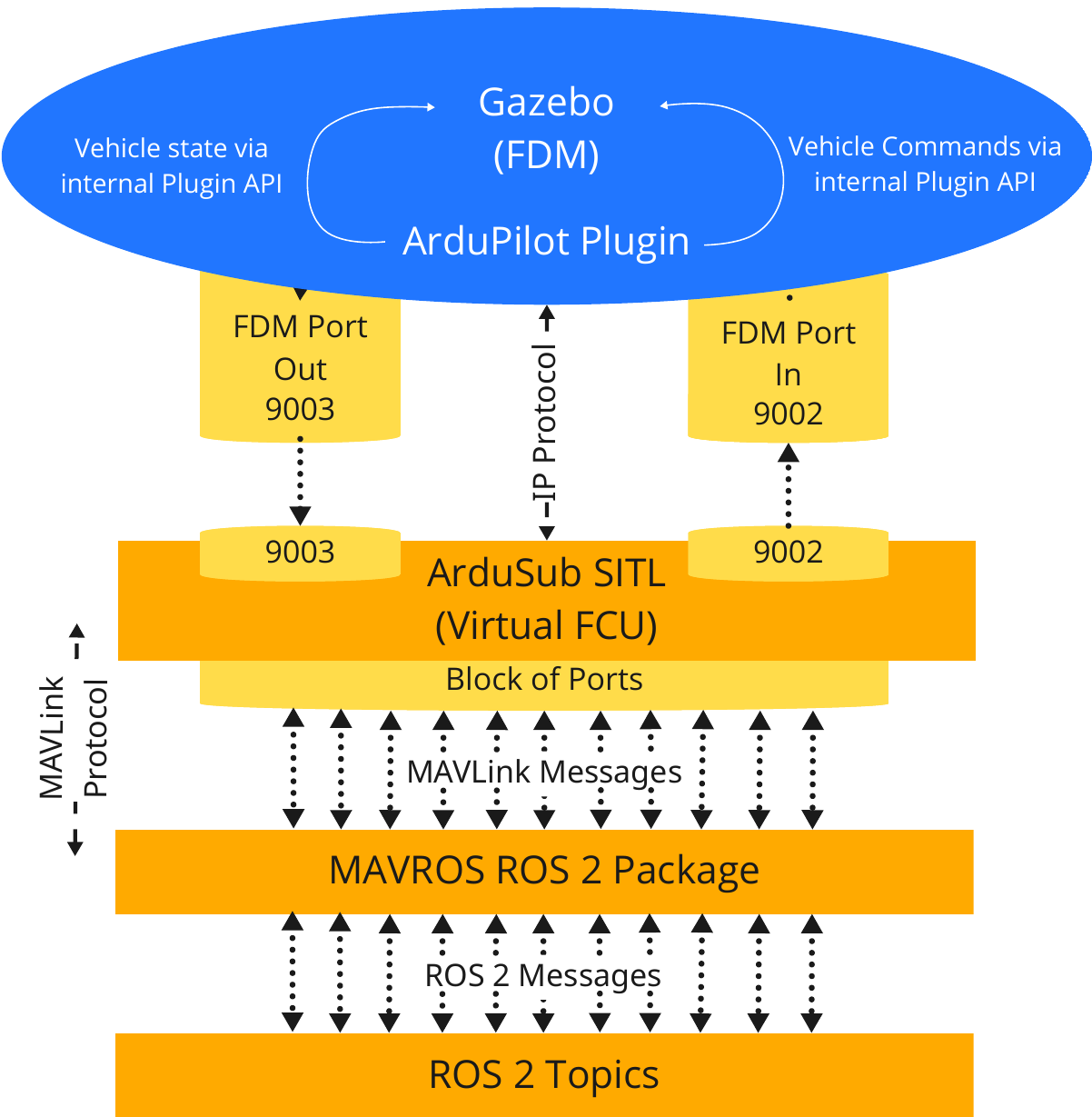}
  \caption{The existing software architecture for simulation in \emph{Angler}.}
  \label{figure:angler_arch}%
\end{figure}

There is widespread use of simulation in underwater robotics~\cite{zebin2024urobench}, and there has been focus in developing more accurate simulation to simulate aquatic conditions, e.g.,~\cite{bingham2019toward}. However, there are few underwater simulators that support multi-robot systems, summarized in~\cref{tab:simulator_comparison}. Stonefish~\cite{grimaldi2025stonefish} is a Bullet-based~\cite{coumans2015bullet} simulator capable of simulating multi-robot systems. It utilizes the collision detection and multi-body dynamics modeling functionalities of Bullet at its core. However, it is purely a simulator and does not offer any provision for hardware deployment. The UVMS Simulator~\cite{edward2025underwater} on the other hand, is designed as a hardware deployment framework. It supports hardware-in-the-loop testing, allowing it to directly interface with a BlueROV Heavy robot and perform tests on hardware. However, both of these simulators lack ArduSub simulation, which is critical for a hardware-software integration pipeline that is closer to real-world systems without requiring actual hardware. 

DAVE~\cite{mabel2022dave} and the UUV Simulator~\cite{musa2016uuv} are other widely used underwater simulators built on top of Gazebo. While DAVE supports multi-robot simulation, it is built with ROS~1. With the release of ROS~2~\cite{macenski2022robot}, the older ROS framework will be deprecated by the end of 2025. Moreover, ROS~2 offers better support for real-world deployments along with the core functionalities of ROS. To exploit the newer features of ROS~2 and to prevent them from becoming obsolete, it is advisable to build systems using ROS~2.
Although efforts are being made to develop DAVE using ROS~2, it has not yet been officially released~\cite{mabel2022dave}.

Our decision to extend \emph{Angler}~\cite{evan2024angler} rather than other existing simulators stems from several considerations. \emph{Angler} is similar to the Marine Vehicle Packages project~\cite{emir2022working}, which provides a framework for control and high-level autonomy. However MVP is strictly targeted towards single-system deployments.
\emph{Angler} has a modular architecture that enabled us to modify and extend individual modules to support multi-robot systems. Our main objective is to provide a platform that allows testing integrated multi-robot motion planning algorithms. \emph{Angler} abstracts the robot-specific controller stack using cascade control architecture, allowing the development of capabilities to support multi-robot motion planning. Additionally, its ability to interface with ArduSub SITL~\cite{ArduPilot} is of utmost importance to us, as it enables us to test multi-robot software architecture in simulation before actual deployment.

\begin{table}[]
    \centering
    \begin{tabularx}{\columnwidth}{r|c|c|c|c}
        &&&\textbf{ArduSub} & \textbf{Cascade} \\
        \textbf{Frameworks} & \textbf{Simulator} & \textbf{ROS~2} & \textbf{SITL} & \textbf{Control} \\
        \hline
        MR Angler (Ours) & Gazebo & \checkmark & \checkmark & \checkmark \\
        UVMS Simulator~\cite{edward2025underwater} & Gazebo & \checkmark & $\times$ & $\times$ \\
        Stonefish~\cite{grimaldi2025stonefish} & Bullet & \checkmark & $\times$ & $\times$ \\
        DAVE~\cite{mabel2022dave} & Gazebo & $\checkmark$ & $\times$ & $\times$ \\
        HoloOcean~\cite{easton2022holoocean} & Unreal & \checkmark & $\times$ & $\times$ \\
        LRAUV~\cite{timothy2023from} & Gazebo & $\checkmark$ & $\times$ & $\times$ \\
        UUV Simulator~\cite{musa2016uuv} & Gazebo* & $\times$ & $\times$ & $\times$ \\
        $ds\_sim$~\cite{ian2025dssim} & Gazebo* & $\times$ & $\times$ & $\times$ \\
        UWSim-NET~\cite{centelles2019uwsimnet} & Gazebo* & $\times$ & \checkmark & $\times$ \\
        \hline
    \end{tabularx}
    \caption{Comparison of multi-robot simulation capabilities among existing frameworks that support multi-robot systems. Gazebo* indicates Gazebo Classic in ROS~1.}
    \label{tab:simulator_comparison}
\end{table}

\subsection{Angler Software Architecture} \label{sec:angler_arch}

\emph{Angler} is an open-source software framework designed for developing autonomous capabilities for underwater robots. For simulation purposes, it interfaces with Gazebo as the flight dynamic model (FDM) that simulates the robot and onboard sensors. An ArduPilot~\cite{ArduPilot} plugin establishes a two-way bridge between the simulator and the ArduSub SITL to exchange data and actuator commands.

\begin{figure*}[t]
  \centering
  \includegraphics[width=17cm,trim={0cm 0cm 0cm 0cm},clip]{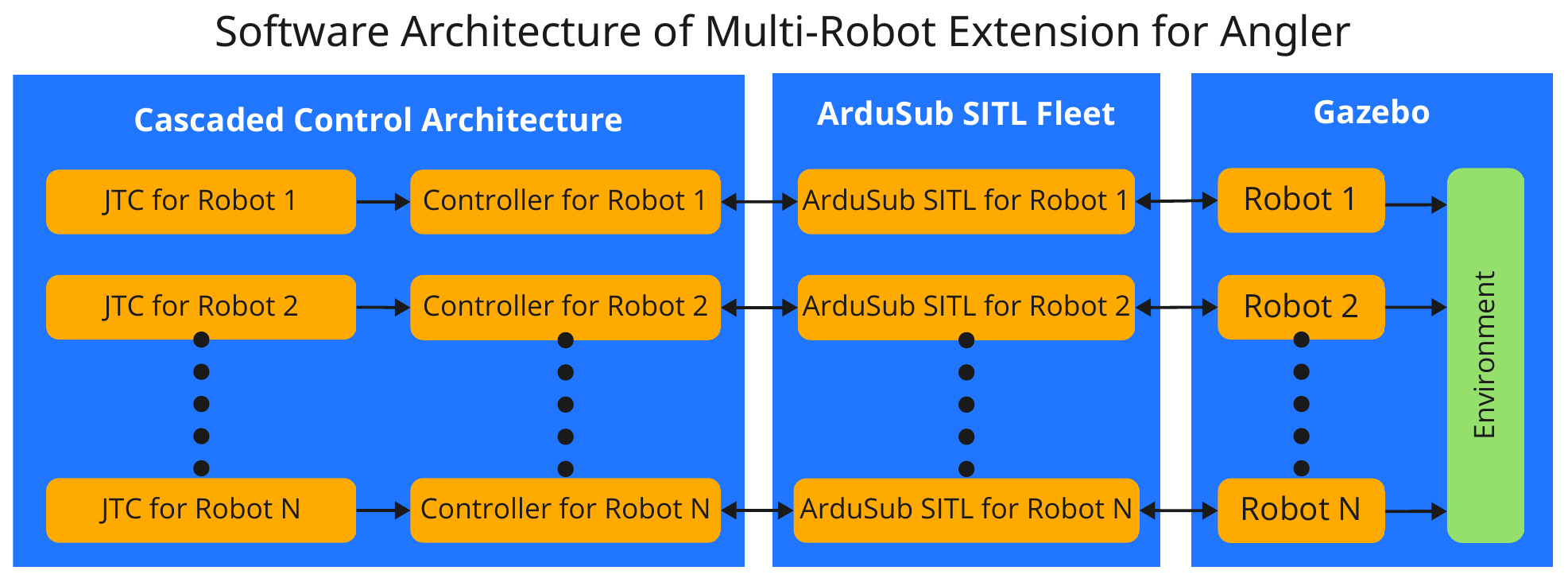}
  \caption{Software architecture illustrating our extension of \emph{Angler}~(\cref{figure:angler_arch}) to enable multi-robot simulation.}
  \label{figure:sim_arch}%
\end{figure*}

\emph{ArduSub} is an autopilot system for underwater vehicles from ArduPilot project.
Ardusub SITL functions as a virtual flight control unit (FCU) that simulates the behavior of underwater systems FCUs, such as Navigator\footnote{\url{https://bluerobotics.com/store/comm-control-power/control/navigator/}}. The FCU communicates with the Ground Control System (GCS) running on ground computers over UDP/TCP connections following the MAVLink Protocol. 
While ROS~2 serves as the de facto software framework for developing robotic applications, it lacks native support for MAVLink messages. To address this limitation, MAVROS---a ROS~2 package---acts as a proxy GCS that decodes MAVLink messages and publishes them as ROS messages on ROS Nodes. To accurately replicate real hardware-software integration, ArduSub SITL interfaces with Gazebo using IP protocol and with ROS~2 through MAVROS. The entire \emph{Angler} architecture is given in a diagram in~\cref{figure:angler_arch}.

However, initially, \emph{Angler} was developed for a single robot use case, leading to hard-coded configurations for interfacing between subsystems, which constrains multi-robot simulation capabilities.

%% file: files/03-simulation-features.tex
\section{Underwater Multi-Robot Simulation}

Our simulation framework is designed for research and development of solutions for multi-robot underwater applications including exploration and transportation tasks. We offer integrated support for autopilot simulation which enables testing the software pipeline along with planning and control algorithms before deployment on hardware. We provide an illustration of our software framework in~\cref{figure:sim_arch}. Our framework is informed by the following design goals to support multi-robot simulation:

\begin{enumerate}
    \item \emph{Leverage Existing Features of Angler:} We build our multi-robot simulation capabilities on top of Angler's useful underwater simulation features, including current simulation and vehicle dynamics modeling.
    \item \emph{ROS~2 and Gazebo Integration:} We support the latest ROS~2 and Gazebo versions, to enable building on and integrating newer features such as cascade control.
    \item \emph{Modularity for Multiple Robots:} We extend Angler with multi-robot simulation such that it retains its modularity. For example, we spawn a separate controller manager for each robot to avoid conflicts. We make the interface between different modules such as Gazebo and ArduSub SITL configurable. 
\end{enumerate}

\subsection{Simulation in Gazebo}

We use \emph{Angler}'s Gazebo-based simulation framework for modeling underwater environments. 
It simulates ocean currents and sensor noise, facilitating \emph{in silico} testing of planning and control algorithms under different conditions. 
Our extension to Angler simulates multiple underwater vehicles concurrently, where robot-specific properties, such as added mass, can be set individually using configuration files, allowing a heterogeneous fleet of robots.

To mimic a setup close to real-world deployment, Gazebo interfaces with the Ardupilot plugin to exchange information with other components of the framework. 
To achieve modularity for a multi-robot system, we have configurable Flight Dynamics Model (FDM) ports. This plugin serves as an interface that connects the Gazebo sensor simulation data and actuator commands to the Ardusub Software-In-The-Loop (SITL), which virtualizes the Flight Control Unit (FCU). The bidirectional communication enables execution commands from SITL to be accurately reflected in the Gazebo simulation environment, providing a closed-loop testing capability for complex underwater operations.

\subsection{ArduSub SITL Integration}
As discussed in~\cref{sec:angler_arch}, ArduSub Software-In-The-Loop (SITL) is a firmware that virtualizes the autopilot running on an underwater robot. 
Integrating ArduSub SITL into our testing framework enables us to test on actual vehicle firmware and validate our software pipeline. 
The ArduSub SITL module implements sensor fusion using Extended Kalman Filter (EKF), thus providing realistic localization. 
It is a sandwich module that interacts with the simulator through Flight Dynamics Model (FDM) ports on one side and MAVROS on the other. 
To establish a multi-robot interface channel with MAVROS, each robot requires access to a unique set of ports; additionally, ArduSub SITL generates MAVLink messages following the MAVLink protocol to mimic a real-world communication network. 
We provide easy configuration of distinct MAVLink ports for multiple robots by allocating non-conflicting ports to each robot, thus enabling the simultaneous simulation of multiple ArduSub SITLs for multi-robot operations.

\subsection{Controllers}
\emph{Angler} implements a cascaded control architecture. 
When controlling multiple robots, each robot requires its own set of cascaded controllers. 
One of the issues that prevented multi-robot simulation in other frameworks is name collisions, due to hard-coded assumptions in controllers. 
We have a modular framework that automatically sets the prefixes for \texttt{tf2} frames and correctly namespaces the nodes and topics. 
However, controller stack is managed through controller managers. 
We spawn robot-specific controller managers to track the controller fleet along with the corresponding hardware interfaces. 

Many robotic applications require a combination of autonomous and teleoperated control. 
For instance, a motion planning algorithm might navigate a robot to the vicinity of a sample collection point, but final precision manipulation may require human oversight.
We integrate with the \texttt{teleop\_twist\_keyboard} package to provide simultaneous teleoperation capabilities for multiple robots.

%% file: files/04-multi-robot-motion-planning.tex
\section{Multi-Robot Motion Planning}

\begin{figure}[t]
  \centering
  \includegraphics[width=8cm,trim={0cm 0cm 0cm 0cm},clip]{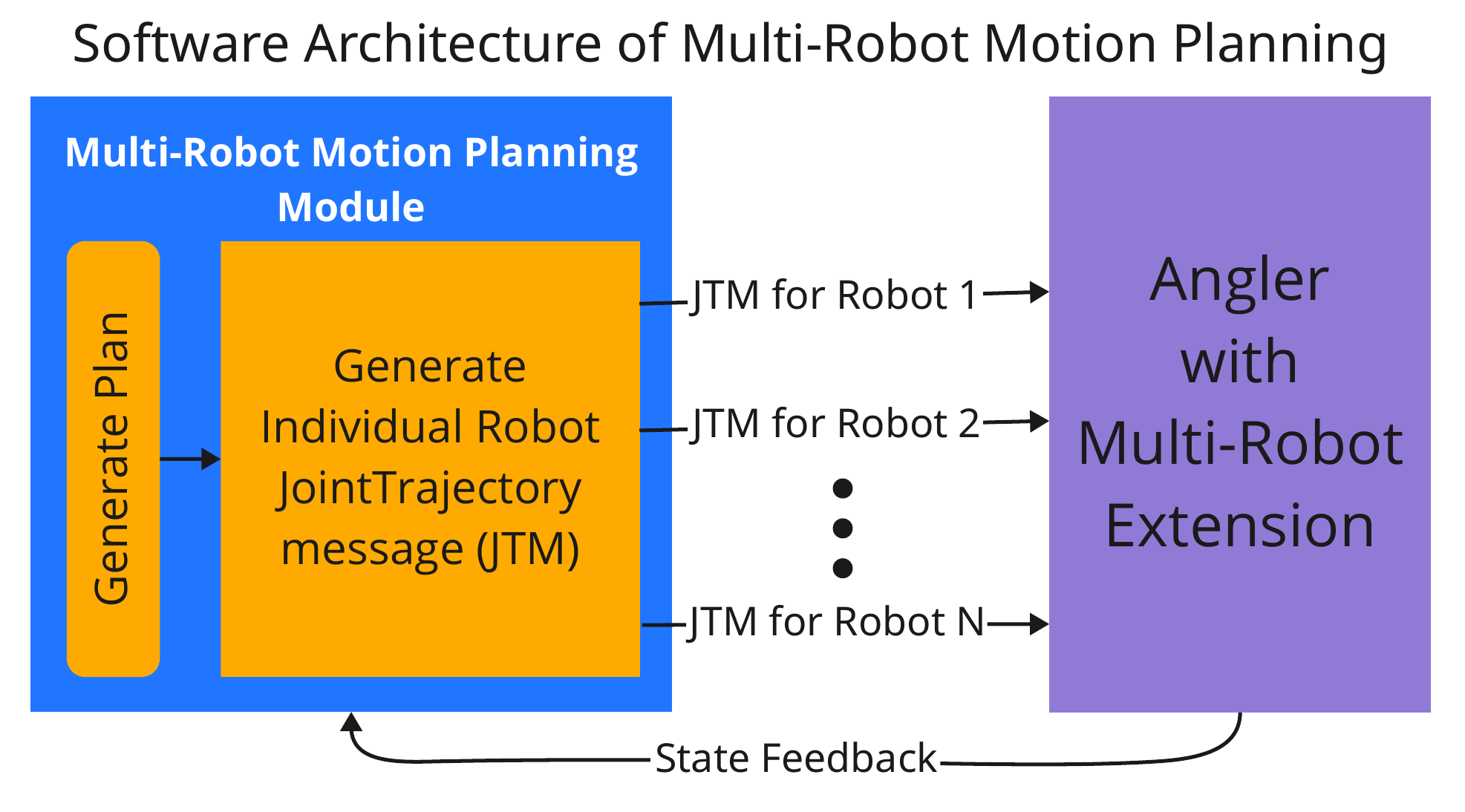}
  \caption{Software architecture for multi-robot motion planning.}
  \label{figure:mmmp_arch}%
\end{figure}

We design a multi-robot motion planning framework that interfaces with the cascaded control architecture of \emph{Angler} via a \texttt{JointTrajectory} controller, shown in~\cref{figure:mmmp_arch}. 
This enables execution of motion plans within the Gazebo simulation environment. 
Our planning framework has the ability to receive state feedback from \emph{Angler} which can then be processed and used for updating the motion plans.

\subsection{JointTrajectory Controller Integration}

To connect the motion planner and cascaded controller stack from \emph{Angler}, we use  \texttt{ros2\_control's} \texttt{JointTrajectory} controller as the preceding controller. 
The \texttt{JointTrajectory} controller  accepts \texttt{JointTrajectory} messages from the motion planner, ensuring planner-controller independence through a standard interface. 
This modularity enables the motion planner to remain agnostic to the underlying control implementation while supporting arbitrary controller stacks that can be integrated with \texttt{JointTrajectory} controller.

\subsection{Collision Avoidance Module}

Collision checking is empirically one of the most computationally expensive operations in motion planning~\cite{joshua2011massively}, with complexity increasing proportionally to the number of bodies in the environment. 
Although Gazebo offers a plugin for collision detection, we need a fast method to check for collisions of any potential state of the system to generate plans. 
FCL~\cite{pan2012fcl}, a popular C++ library that provides fast collision checking, is used within numerous frameworks, e.g., MoveIt, Trimesh, and Pinocchio~\cite{moveit,trimesh,pinocchio}.
In particular,  
we are interested in exploring cooperative construction activities underwater, which can utilize manifolds for defining constraints~\cite{agrawal2024constrained}, which is well supported by Pinocchio. 
Hence, we chose Pinocchio to develop a multi-robot collision checking module. 
This module allows us to check for inter-robot collisions, along with external collisions with the environment.

\subsection{OMPL Integration}

The Open Motion Planning Library (OMPL)~\cite{sucan2012the} offers a rich collection of state-of-the-art sampling-based motion planners.
We integrate OMPL with our multi-robot collision checking and modeling to generate trajectories for the entire multi-robot system~\cref{figure:mmmp_arch}.
Here, we model the multi-robot system as a composition of the individual configuration spaces, planning for the motions of the robots simultaneously.
We plan to develop custom OMPL-based motion planners for more complex use cases like collaborative underwater transportation~\cite{agrawal2024constrained}.

\begin{figure}[h]
  \centering
  \includegraphics[width=8.5cm,trim={0cm 0cm 0cm 0cm},clip]{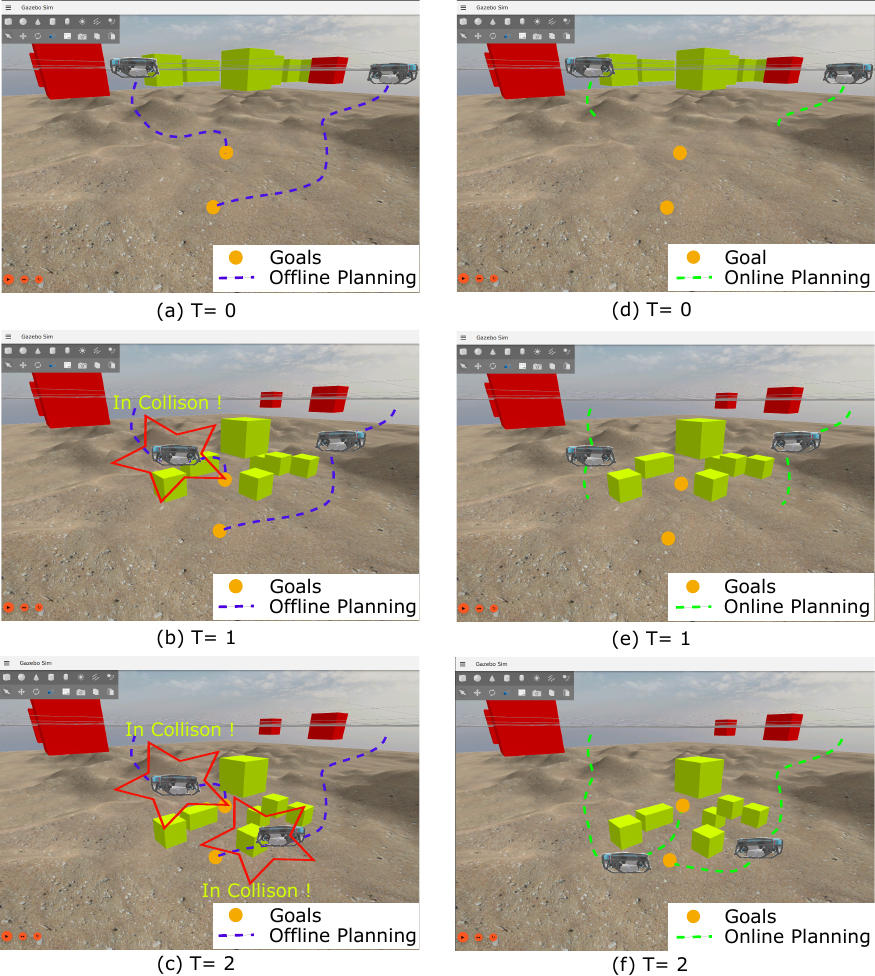}
  \caption{A scene in \emph{Angler} showcasing online and offline motion planning in presence of static and dynamic obstacles for two robots. $T= 0, 1$, and $2$ represents three distinct timesteps.}
  \label{fig:paths}%
\end{figure}

\subsection{Environment Generation Tool}
To validate the robustness of a motion planning algorithm or to identify potential failure cases, we must perform rigorous testing in diverse environments.
For learning-based planners that require datasets to train their models, manually generating environment configurations is impractical and time-consuming. 
We provide a procedural generation approach based on a cellular automata~\cite{cellularAutomata} to populate obstacles in the environment, creating a combination of fixed and dynamic obstacles, an example of which is shown in~\cref{fig:paths}.
Dynamic obstacles represent objects set in motion by ocean currents and the effects of buoyancy, introducing temporal complexity to the navigation problem.
Our tool can generate environments with varying levels of complexity.
Environment complexity is defined by obstacle density and the number of dynamic obstacles.

\subsection{Benchmarking Planning for Underwater Applications}

Our simulation framework and motion planning tools enables multi-robot motion planning and benchmarking of novel motion planners against existing ones, particularly those supported by OMPL. 
We enable the comparison of different algorithms under identical environments and simulation setups using metrics such as computation time, execution time, and success rate.
Our procedural environment generation tool enables us to benchmark the performance of different algorithms in various settings, including environments with static and moving obstacles. 
We can identify failure points of unsuccessful plans by utilizing Gazebo's collision detection capabilities. 
The simulator enables us to find exact parameters that optimizes firmware performance. 
Our framework also supports developing integrated motion planning algorithms that incorporate real-time feedback about robot and obstacle states, thus enabling us to develop and validate multi-robot motion planning algorithms for underwater. 

%% file: files/05-conclusion-and-future-works.tex
\section{Discussion and Conclusion}
We have presented an extension to \emph{Angler} to support simulation of multiple robots, a framework that enables the development and validation of solutions for autonomous multi-robot operations, including sensing, control, and motion planning. The core capabilities that our extension provides over the existing \emph{Angler} framework include establishing non-conflicting communication channels between Gazebo, ArduSub SITL, and MAVROS for each individual robot in a multi-robot setup, where multiple robots can operate in the same environment simultaneously.

We also developed a multi-robot motion planning tool by providing a high-levl \texttt{JointTrajectory} controller that interfaces with robot-specific controller stacks. Our ROS~2 package supports any controller stack that can be interfaced with the \texttt{JointTrajectory} controller. Our OMPL integration and environment generation tool enable rapid development and testing against existing algorithms.

One challenge with the underwater domain is testing for scalability with an increasing number of robots, as real-world testing is expensive. Our framework can be used to estimate the number of robots required for a task. 
We are also interested to evaluate whether our framework can be used for generating datasets or simulating such applications for learning-based methods.

Our ongoing goal is to perform underwater cooperative construction and transportation tasks. Future work will focus on multi-robot mobile manipulation, underwater localization, and task allocation mechanisms for underwater multi-robot operations in marine environments.